\documentclass{ecai}
\usepackage{times}
\usepackage{graphicx}
\usepackage{latexsym}
\usepackage{amsmath}
\usepackage{amssymb}
\usepackage{booktabs}
\ecaisubmission
\usepackage{floatrow}
\usepackage{blindtext}

\usepackage[pagebackref=true,breaklinks=true,colorlinks,bookmarks=false]{hyperref}
\usepackage[nameinlink, noabbrev]{cleveref}
\usepackage[utf8]{inputenc} \usepackage[T1]{fontenc}

\newcommand{\commentMR}[1]{#1}
\newcommand{\commentHG}[1]{#1}
\newcommand{\commentRC}[1]{#1}

\usepackage{tikz}
\usetikzlibrary{shapes.geometric, arrows}
\usetikzlibrary{decorations.pathreplacing}
\tikzstyle{arrow} = [thick,->,>=stealth]
\usetikzlibrary{shadows}

\newcommand{\argmax}[1]{\underset{#1}{\operatorname{argmax}}\;}

\begin{document}

\tikzset{%
  cascaded/.style = {%
    general shadow = {%
      shadow scale = 1, shadow xshift = -1ex, shadow yshift = 1ex, draw, thick, fill = white},
    general shadow = {%
      shadow scale = 1, shadow xshift = -.5ex, shadow yshift = .5ex, draw, thick, fill = white},
    fill = white, draw, thick, minimum width = 1.5cm, minimum height = 0.75cm}}

\title{MetaFusion: Controlled False-Negative Reduction of Minority Classes in Semantic Segmentation}


\author{Robin Chan\institute{University of Wuppertal, School of Mathematics and Natural Sciences, ~Germany, ~ email: \{rchan, rottmann, hgottsch\}@uni-wuppertal.de} \and
Matthias Rottmann\footnotemark[1] \and
Fabian H\"uger\institute{Volkswagen Group Innovation, Center of Innovation Automation, Germany, email: \{fabian.hueger, peter.schlicht\}@volkswagen.de} \and
Peter Schlicht\footnotemark[2] \and
Hanno Gottschalk\footnotemark[1] }

\maketitle
\bibliographystyle{ecai}

\begin{abstract}
\commentHG{In semantic segmentation datasets, classes of high importance are oftentimes underrepresented, e.g., humans in street scenes. Neural networks are usually trained to reduce the overall number of errors, attaching identical loss to errors of all kinds. However, this is not necessarily aligned with human intuition. For instance, an overlooked pedestrian seems more severe than an incorrectly detected one. One possible remedy is to deploy different decision rules by introducing class priors which assign larger weight to underrepresented classes. While reducing the false-negatives of the underrepresented class, at the same time this leads to a considerable increase of false-positive indications.
In this work, we combine decision rules with methods for false-positive detection. We therefore fuse false-negative detection with uncertainty based false-positive meta classification.  We present proof-of-concept results for CIFAR-10, and prove the efficiency of our method for the semantic segmentation of street scenes on the Cityscapes dataset based on predicted instances of the 'human' class. In the latter we employ an advanced false-positive detection method using uncertainty measures aggregated over instances. We thereby achieve improved trade-offs between false-negative and false-positive samples of the underrepresented classes.} 
\end{abstract}

\section{INTRODUCTION}
Deep learning
has improved the state-of-the-art in a broad field of applications such as computer vision, speech recognition and natural language processing by introducing deep convolutional neural networks (CNNs). Although class imbalance is a well-known problem of traditional machine learning models, little work has been done to examine and handle the effects on deep learning models\commentHG{, see however \cite{Johnson2019} for a recent review}. Class imbalance in a dataset occurs when at least one class contains significantly less examples than another class. The performance of CNNs for classification problems has empirically been shown to be detrimentally affected when applied on skewed training data~\cite{Buda2018, Lopez2013} by revealing a bias towards the overrepresented class. 
Being an 
classification problem at 
pixel-level, semantic segmentation therefore is exhibited to the same set of problems when class imbalance is present. As the imbalance naturally exist in most datasets for ``real world'' applications, finding the underrepresented class is of highest interest. 

Methods for handling class imbalance have been developed and can be divided into two main categories: \textit{sampling-based} and \textit{algorithm-based} techniques~\cite{Buda2018, Johnson2019, Krawczyk2016}. While sampling-based methods operate directly on a dataset with the aim to balance its class distribution, algorithm-based methods include a cost schem\commentMR{e} to modify the learning process or decision making of a classifier.

In the simplest form, balancing data is done by randomly discarding samples from frequent (majority) groups and/or randomly duplicating samples from less frequent (minority) groups. These techniques are known as oversampling and undersampling~\cite{VanHulse2007}, respectively. They \commentMR{can} lead to performance improvement, in particular with random oversampling~\cite{Buda2018, Lopez2013, Masko2015} unless there is no overfitting~\cite{Chawla2004}. A more advanced approach called SMOTE~\cite{Chawla2002} alleviates the latter issue by creating synthetic examples of minority classes.

Oversampling methods are difficult to apply on semantic segmentation datasets due to naturally occurring class frequencies on single input frames. Considering the Cityscapes~\cite{Cordts2016Cityscapes} dataset of urban street scenes for instance, the number of annotated road pixels exceeds the number of annotated person pixels by a factor of \commentRC{roughly 25 despite the fact that persons already are strongly represented in this datatset} as street scenarios are shown from a car driver's perspective. 

The training approach is to assign costs to different classification mistakes for different classes and include them in the loss function~\cite{Bulo2017, Caesar2015, Wang2016}. Instead of minimizing the total error, the \commentRC{average} misclassifcation cost is minimized. In addition, methods have been proposed learning the cost parameters throughout training~\cite{Khan2018, Zhang2016} and thus eliminating the ethical problem of predefining them~\cite{Chan2019Dilemma}. These methods require only little tuning and outperform sampling-based approaches without significantly affecting training time. 
Modifying the loss function however biases the CNN's output.

One approach to correct class imbalance during inference is output thresholding, thus interchanging the standard maximum a-posteriori probability (MAP) principle for an alternate decision rule.
Dividing the CNN's output by the estimated prior probabilities for each class was proposed in~\cite{Buda2018, Chan2019} which is also known as Maximum Likelihood rule in decision theory~\cite{Fahrmeir1996}.
This results in a reduced likelihood of misclassifying minority class objects and a performance gain in particular with respect to the sensitivity of rare classes. Output thresholding does neither affect training time nor the model's capability to discriminate between different groups. It is still a suitable technique for reducing class bias as it shifts the priority to predicting certain classes that can be easily added on top of every CNN.

In the field of semantic segmentation of street scenes the overall performance metric intersection over union (IoU)~\cite{Everingham2015} is mainly used. This metric is highly biased towards large and therefore majority class objects such as street or buildings. 
\commentMR{Currently,} state-of-the-art models
achieve class IoU scores of 83\% for Cityscapes~\cite{Cordts2016Cityscapes} and 73\% for Kitti~\cite{Geiger2013Kitti}.
Further maximizing global performance measures is important but does not necessarily improve the overall system performance. The priority shifts to rare and potentially more important classes, where the lack of reliable detection has potentially fatal consequences in applications like automated driving. 

In this context, uncertainty estimates are helpful as they can be used to quantify how likely an incorrect prediction has been made. Using the maximum softmax probability as confidence estimate has been shown to effectively identify misclassifications in image classification problems which can serve as baseline across many other applications~\cite{Hendrycks2017}. More advanced techniques include Bayesian neural networks (BNNs) that are supposed to output distributions over the model's weight parameters~\cite{neal2012}. As BNNs come with a prohibitive computational cost, recent work\commentMR{s} developed approximations such as Monte-Carlo dropout~\cite{gal2016} or stochastic batch normalization~\cite{Atanov2019}. These methods generate uncertainty estimates by sampling, i.e., through multiple forward passes. These sampling approaches are applicable for most CNNs as they do not assume any specific network architecture, but they tend to be computationally expensive during inference. Other frameworks include learning uncertainty estimates via a separate output branch in CNNs~\cite{devries2018learning,Kendall2017} which seems to be more appropriate in terms of computational efficiency for real-time inference. 

In semantic segmentation, uncertainty estimates are usually visualized as spatial heatmaps. 
Nevertheless, it is possible that CNNs \commentMR{show poor performance but also high confidence scores}
\cite{Amode2016}. 
Therefore, auxiliary machine learning models for predicting the segmentation quality~\cite{Kohlberger2012,Zhang2016} have been proposed. While some methods built upon hand-crafted features, some other methods apply CNNs for that task by learning a mapping from the final segmentation to its prediction quality~\cite{DeVries2018,Huang2016}. 
A segment-based prediction rating method for semantic segmentation was proposed in~\cite{rottmann2018} and extended in~\cite{rottmann2019,Maag2019}. They derive  aggregated dispersion metrics from the CNN's softmax output and pass them through a classifier that discriminates whether one segment intersects with the ground truth or not. These hand-crafted metrics have shown to be well-correlated to the calculated segment-wise IoU. The method is termed ``\emph{MetaSeg}'' which we use from now on to refer to that procedure.

In this work, we present a novel method for semantic segmentation in order to reduce the false-negative \commentMR{rate} of rare class objects and alleviate the effects of strong class imbalance in data.
The proposed method consists of two steps: First, we apply the Maximum Likelihood decision rule that adjusts the neural network's probabilistic / softmax output with the prior class distribution estimated from the training set. In this way, less instances of rare classes are overlooked but to the detriment of producing more false-positive predictions of the same class. Afterwards, we apply MetaSeg to extract dispersion measures from the balanced softmax output and, based upon that, discard the additional false-positive segments in the generated segmentation mask.

This work mainly builds on methods already presented in~\cite{Chan2019} and~\cite{rottmann2018}. Our main contribution is the fusion of these two components providing an additional 
segmentation mask that is \commentRC{more} sensitive to finding rare class objects, but keeps false-positive instances in check. 
\commentRC{
Compared to different class weightings for decision thresholding, we obtain a more favorable trade-off between error rates.}
As inference post-processing tool, our method does not touch the underlying CNN architecture used for semantic segmentation, it is computationally cheap, easily interpretable and can be seamlessly added on top of other CNNs for semantic segmentation.

\begin{figure*}
\centerline{\scalebox{1.0}{\begin{tikzpicture}[node distance=3cm]

\node (sf) [align=center,cascaded] at (0,0) {};
\node (sfp) [align=center,cascaded] at (0,-2) {};
\node (bay) [draw, thick, minimum width = 1.5cm, minimum height = .75cm] at (3,0) {};
\node (ml) [draw, thick, minimum width = 1.5cm, minimum height = .75cm] at (3,-2) {};
\node [diamond, aspect=2] (metaseg) [draw, thick] at (6.5,-1) {\small predict};
\node (kick) [ellipse, draw, thick, align=center] at (9,0) {$= 0$};
\node (keep) [ellipse, draw, thick, align=center] at (9,-2) {$> 0$};
\node (out) [draw, thick, minimum width = 1.5cm, minimum height = .75cm] at (12,-1) {};

\draw [arrow] (sf) -- (sfp) node [midway, fill=white] {\scriptsize adjust with priors};
\draw [arrow] (sf) -- (bay) node [midway,above] {\scriptsize $\mathop{argmax}$};
\draw [arrow] (sfp) -- (ml) node [midway,above] {\scriptsize $\mathop{argmax}$};
\draw [arrow] (sfp) to [out=270,in=270] (metaseg);
\draw [arrow] (ml) -- (3,-1) -- (metaseg) node [midway,above] {\scriptsize ML segments with};
\draw [arrow] (bay) -- (3,-1) -- (metaseg) node [midway,below] {\scriptsize $\textit{IoU(ML,Bayes)} = 0$};
\draw [arrow] (metaseg) to [out=0,in=180] (kick);
\draw [arrow] (metaseg) to [out=0,in=180] (keep);

\draw [arrow] (kick) to [out=0,in=180] (out) node [near end, above] {};
\draw [arrow] (keep) to [out=0,in=180] (out) node [near end, above] {};

\node [align=center, text width = 2cm, fill = white] at (0,1.3) {Softmax Probabilities};
\node [align=center, text width = 2cm, fill = white] at (0,-3.1) {Balanced Probabilities};
\node [align=center, text width = 2cm] at (3,1.05) {Bayes mask};
\node [align=center, text width = 2cm] at (3,-2.9) {ML mask};
\node [align=center, text width = 2cm] at (6.5,1.08) {MetaSeg};
\node [align=center, text width = 2cm] at (9.2,1.08) {$\textit{IoU(ML,GT)}$};
\node [fill=white, align=center] at (6.45,-2.2) {\scriptsize derive uncertainty metrics};
\node [fill=white, align=center] at (10.4,-0.45) {\scriptsize believe Bayes};
\node [fill=white, align=center] at (10.4,-1.55) {\scriptsize believe ML};
\node [align=center] at (12,1.05) {Output};

\end{tikzpicture}}}
\caption{Overview of our method for controlled false-negative reduction of minority classes which we term ``\emph{MetaFusion}''. Note that \textit{IoU} denotes the intersection over union measure of two segmentation masks.} \label{fig:metafusion-overview}
\end{figure*}
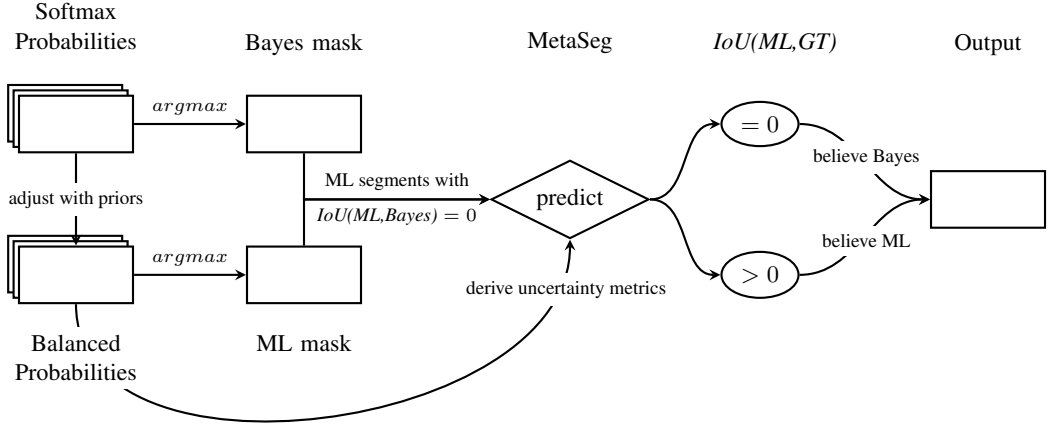

\commentRC{This work is structured as follows: In sections~\ref{sec:ml} and~\ref{sec:prediction-error-classification}, we recall the building blocks of our approach, namely the Maximum Likelihood decision rule for the reduction false-negatives and MetaSeg for false-positive segments detection, respectively. In~\cref{sec:metafusion}, we combine the latter components and show proof-of-concept results for CIFAR-10 in~\cref{sec:cifar10}. We complement this work by extending our approach to the application-relevant task of semantic segmentation and show numerical results for the Cityscapes data in~\cref{sec:cityscapes}.}

\section{MAXIMUM LIKELIHOOD DECISION RULE}\label{sec:ml}
Neural Networks for semantic segmentation can be viewed as statistical models providing pixel-wise probability distributions that express the confidence of predicting the correct class label $y$ within a set $\mathcal{Y} := \{1,\ldots,l\}$ of predefined classes. The classification at pixel location $z \in \mathcal{Z}$ is then performed by applying the $\mathop{argmax}$ function to the posterior probabilities / softmax output $p_z(y|x) \in [0,1]$ after processing image $x \in \mathcal{X}$. In the field of Deep Learning, this decision principle, called the maximum a-posteriori probability (MAP) principle, is by far the most commonly used one:
\begin{align}
    d_{\mathit{Bayes}}(x)_z := \argmax{y\in\mathcal{Y}} p_z(y|x) ~ .
\end{align}
In this way, the overall risk of incorrect \commentMR{classifications} 
is minimized, i.e., for any other decision rule $d: [0,1]^{|\mathcal{Z}|} \mapsto \mathcal{Y}^{|\mathcal{Z}|}$ and with
\begin{align}\label{eq:sim-risk}
    R_\mathit{sym}(d) := \frac{1}{|\mathcal{Z}|} \sum_{z \in Z} \sum_{y \in Y} 1_{\{ d(x)_z \neq y \}} p_z(y|x) ~ \forall ~ x \in \mathcal{X}
\end{align}
it holds $R_\mathit{sym}(d_{\mathit{Bayes}}) \leq R_\mathit{sym}(d)$.
In decision theory, this principle is also known as Bayes decision rule~\cite{Fahrmeir1996} and it incorporates knowledge about the prior class distribution $p(y)$. As a consequence, in cases of large prediction uncertainty the MAP / Bayes rule tends to predict classes that appear frequently in the training dataset when applied in combination with CNNs. However, classes of high interest might appear less frequently.
Regarding highly unbalanced datasets the Maximum Likelihood (ML) decision rule \commentMR{oftentimes is a good choice as it compensates for the weights of classes induced by priors:} 
\begin{align}\label{eq:ml-rule}
    \hat{y}_z = d_{\mathit{ML}}(x)_z := \argmax{y\in\mathcal{Y}}p_z(x|y) = \argmax{y\in\mathcal{Y}} \frac{p_z(y|x)}{p_z(y)} ~ .
\end{align}
Instead of choosing the class with the largest a-posteriori probability $p_z(y|x)$, the ML rule chooses the class with the largest conditional likelihood $p_z(x|y)$. It is optimal regarding the risk function
\begin{align}
    R_\mathit{inv}(d) := \frac{1}{|\mathcal{Z}|} \sum_{z \in Z} \sum_{y \in Y} 1_{\{ d(x)_z \neq y \}} p_z(x|y) ~ \forall ~ x \in \mathcal{X}
\end{align}
and in particular $R_\mathit{inv}(d_{\mathit{ML}}) \leq R_\mathit{inv}(d_{\mathit{Bayes}})$ is satisfied.
The ML rule corresponds to the Maximum Likelihood parameter estimation in the sense that it aims at finding the distribution that fits best the observation. In our use case, the ML rule chooses the class that is most typical for a given pattern observed in an image independently of any prior belief, such as the frequency, about the semantic classes.
Moreover, the only difference between these two decision rules lies in the adjustment by the priors $p_z(y)$ (see~\cref{eq:ml-rule} and Bayes' theorem~\cite{bayes-theorem}).

Analogously to~\cite{Chan2019}, we approximate $p_z(y)$ in a position-specific manner using the pixel-wise class frequencies of the training set:
\begin{equation}\label{eq:prior-esti}
    \commentRC{
    \hat{p}_z(y) = \frac{1}{|\mathcal{X}|} \sum_{x \in \mathcal{X}} 1_{\{y_z(x) = y\}} ~ \forall ~ y \in \mathcal{Y}, z \in \mathcal{Z} ~ . 
    }
\end{equation}
After applying the ML rule, the amount of overlooked rare class objects is reduced compared to the Bayes rule, but to the detriment of overproducing false-predictions of the same class. Hence, our ultimate goal is to discard \commentRC{as many} 
additionally produced false-positive segments \commentRC{as possible} while keeping \commentRC{almost} all additionally produced true-positive segments.

\section{PREDICTION ERROR CLASSIFICATION}\label{sec:prediction-error-classification}
In order to decide which \commentRC{additional} \commentMR{segments predicted by ML but not by Bayes} to discard in an automated fashion, we train a binary classifier performing on top of the CNN for semantic segmentation analogously to~\cite{rottmann2018, rottmann2019}.
Given the conditional likelihood (softmax output adjusted with priors), we estimate uncertainty per segment by aggregating different pixel-wise dispersion measures, such as entropy
\begin{equation}
    E_z(x) = -\frac{1}{\log(|\mathcal{Y}|)} \sum_{y \in \mathcal{Y}} p_z(x|y) \log(p_z(x|y)) ~ \forall ~ z \in \mathcal{Z},
\end{equation}
probability margin
\begin{equation}
    M_z(x) = 1 - p_z(x|\hat{y}_z) + \max_{y \in \mathcal{Y} \setminus \{ \hat{y}_z \} } p_z(x|y) ~ \forall ~ z \in \mathcal{Z}
\end{equation}
or variation ratio
\begin{equation}
    V_z(x) = 1 - p_z(x|\hat{y}_z) ~ \forall ~ z \in \mathcal{Z} ~ .
\end{equation}
As uncertainty is typically large at transitions from one class to another \commentMR{(in pixel space, i.e., at transitions between different predicted objects)}, we additionally treat these dispersion measures separately for each segment's interior and boundary. The generated uncertainty estimates serve as inputs for the auxiliary ``meta'' model which classifies into the classes $\{\mathit{IoU}=0\}$ and $\{\mathit{IoU}>0\}$. Since the classification is employed on segment-level, the method is also termed ``\emph{MetaSeg}''.

We only add minor modifications to the approach for prediction error classification, in the following abbreviated as ``meta'' classification, compared to~\cite{rottmann2018}. For instance, instead of computing logistic least absolute shrinkage and selection operator (LASSO~\cite{Tibshirani1996}) regression fits, we use gradient-boosting trees (GB~\cite{hastie2009}). GB has shown to be a powerful classifier on binary classification problems and structured data with modest dataset size which both match our problem setting.

\commentRC{Additionally to the uncertainty measures, we introduce further metrics indicating incorrect predictions.}
For localization purposes we include a segment's geometric center
\begin{equation}
    G_h(k) = \frac{1}{|k|} \sum_{i=1}^{|k|} h_i ~ , ~ G_v(k) = \frac{1}{|k|} \sum_{j=1}^{|k|} v_j
\end{equation}
with $k = \{ ( h_s, v_s ) \in \mathcal{Z}, {s=1,\ldots,|k|}\}\in \hat{\mathcal{K}}_x$ being the pixel coordinates of one segment / connected component in the predicted segmentation mask, i.e., a set consisting of neighboring pixel locations with the same predicted class. 
The geometric center is the mean of all coordinates of a segment in all directions, in our case in horizontal and vertical direction.

Another metric to be included makes use of a segment's surrounding area to determine if an object prediction is misplaced. Let $k_{nb}=\{ (h^\prime,v^\prime) \in [h \pm 1] \times [v \pm 1] \subset \mathcal{Z}:  (h^\prime,v^\prime) \notin k, (h,v) \in k \}$ be the neighborhood of $k \in \mathcal{\hat{K}}_x$. Then, regarding segment $k$,
\begin{equation}
    N(k|y) = \frac{1}{|k_{bd}|} \sum_{z \in k_{bd}} 1_{\{ \hat{y}_z = y \}} ~ \forall ~ y \in \mathcal{Y}
\end{equation}
expresses the ratio of the amount of pixels in the neighborhood predicted to belong to class $y$ to neighborhood size.

\section{COMBINING MAXIMUM LIKELIHOOD RULE AND META CLASSIFICATION}\label{sec:metafusion}

\begin{figure}[t]
\begin{minipage}[b]{1.0\linewidth}
  \centering
  \centerline{\begin{tikzpicture}
    \def\firstellipse{(-1,0) ellipse (2 and 0.5)}
    \def\secondellipse{(1,0) ellipse (1.5 and 0.5)}
	\def\thirdellipse{(-0.7,0) ellipse (1 and 0.4)}
	
    \fill[red!30!white] \firstellipse;
    \fill[blue!30!white]  \secondellipse;
    \fill[yellow!30!white]  \thirdellipse;

    \begin{scope}
        \clip \firstellipse;
        \fill[green!60!white] \secondellipse;
    \end{scope}

    \begin{scope}
        \clip \thirdellipse;
        \fill[green!30!white] \secondellipse;
    \end{scope}

    \draw \firstellipse \secondellipse \thirdellipse;
    
    \node [align=center] at (-2.2,0) {\textbf{ML}};
    \node [align=center] at (-0.9,0) {\textbf{B}};
    \node [align=center] at (1.6,0) {\textbf{GT}};
 
	\node [align=center,text width=1cm] at (3.2,0) {\color{blue!60!white}\textbf{Ground Truth}};
	\node [align=center, text width=1cm] at (-3.8,0) {\color{orange!90!white}\textbf{Predic-tion}};
	
\end{tikzpicture}}
  \caption{Graphical illustration of the relation between Bayes and ML prediction segments for rare classes.}
  \label{fig:bay-ml-rel}
\end{minipage}
\end{figure}
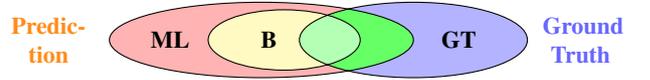

After describing the key components of our method for controlled false-negative reduction in the preceding sections, we now present our approach as combination of \commentMR{the} Maximum Likelihood decision rule and prediction error meta classification for semantic segmentation in more detail. For the most underrepresented class $c \in \mathcal{Y}$ in an unbalanced semantic segmentation dataset,
\commentRC{in many real-world applications often also the class of highest interest,} all predicted Bayes segments are inside ML segments~\cite{Chan2019}, see~\cref{fig:bay-ml-rel}. Consequently, for $c$ we assume that a non-empty intersection between \commentMR{an} ML segment and any Bayes segment (predicted to belong to $c$) indicates a confirmation for the presence of a 
minority class object that was already detected by Bayes. In this case we say \emph{the decision rules agree}. More crucial are predicted ML segments that do not intersect with any Bayes segment of the same class, i.e., \emph{the decision rules disagree}, as these indicate a CNN's uncertain regions where rare instances are potentially overlooked.

\commentMR{The observation} whether the decision rules agree \commentMR{or not} build\commentMR{s} the basis for segment selection for further processing. Let $k \in \hat{\mathcal{K}}_{x,\mathit{ML}}$ be the pixel coordinates of one connected component in the ML mask. Then, given input $x$,
\begin{equation}
    \mathcal{D}_x = \{ k \in \hat{\mathcal{K}}_{x,\mathit{ML}} : d_\mathit{ML}(x)_z \neq d_\mathit{Bayes}(x)_z ~ \forall ~ z \in k \}
\end{equation}
denotes the set of segments in which Bayes and ML disagree. Restricting $\mathcal{D}_x$ 
\commentMR{to} 
\commentMR{a single} minority class $c \in \mathcal{Y}$, 
we obtain the subset $\mathcal{D}_{x|c} \commentRC{=} \{ k_c \in \mathcal{D}_x : d_\mathit{ML}(x)_z = c ~ \forall \, z \in k_c \}$. The obtained subset contains the candidates we \commentMR{process with} MetaSeg. 
\commentRC{Let $\mu_{k} : [0,1]^{|\mathcal{Z}| \times |\mathcal{Y}|} \mapsto \mathbb{R}^q$ be a vector-valued function that returns a vector containing all generated input metrics for MetaSeg restricted to segment $k \in \mathcal{D}_{x|c}$.} 
We derive aggregated uncertainty metrics per segment
\begin{equation}
        U_k := \mu_{k} ((\hat{p}(x|y))_{y \in \mathcal{Y}}) ~ \forall ~ k \in \mathcal{D}_{x|c}
\end{equation}
that serve as input for the meta classifier, see also~\cref{sec:prediction-error-classification} and cf.~\cite{rottmann2018, rottmann2019}.
The classifier we use in our meta model is gradient-boosting tree algorithm (GB~\cite{hastie2009}) and it is trained to discriminate between true-positive and false-positive segment prediction. Thus, we seek a function $\hat{f}: \commentRC{\mathbb{R}}^{q} \mapsto \{ 0,1 \}$ that learns the mapping
\begin{equation}
    f(U_{k}) =
    \begin{cases}
    1, ~ \text{if} ~~ \exists ~ z \in k: d_\mathit{ML}(x)_z = y_z \\
    0, ~ \text{else}
    \end{cases}
\end{equation}
with one connected component $k \in \mathcal{D}_{x|c}$ being considered as true-positive if there exists (at least) one pixel assigned to the correct class label and as false-positive otherwise. In the latter case, we remove that segment from the ML mask and replace it with the Bayes prediction. For the remaining connected components $k^\prime \in \hat{\mathcal{K}}_{x,\mathit{ML} } \setminus \mathcal{D}_{x|c}$, whether or not they are minority class segments, we stick to the Bayes decision rule as well as it is optimal with respect to the expected total number of errors, see~\cref{eq:sim-risk}.
Therefore, the final segmentation output
\begin{equation}
    d_\mathit{Fusion}(x)_z =
    \begin{cases}
        d_\mathit{ML}(x)_z, \, \text{if} ~ \hat{f}(U_{k}) = 1 \land z \in k \in \mathcal{D}_{x|c} \\
        d_\mathit{Bayes}(x)_z, \, \text{else}
    \end{cases}
\end{equation}
fuses Maximum Likelihood and Bayes decision rule. In this way, compared to standard MAP principle, we sacrifice little in overall performance but significantly improve performance on segment recall. We term our approach ``\emph{MetaFusion}'' and provide a summary as graphical illustration in~\cref{fig:metafusion-overview}.

\section{NUMERICAL RESULTS FOR CIFAR-10}\label{sec:cifar10}
\begin{figure}[t]
    \centering
    \includegraphics[width=\linewidth]{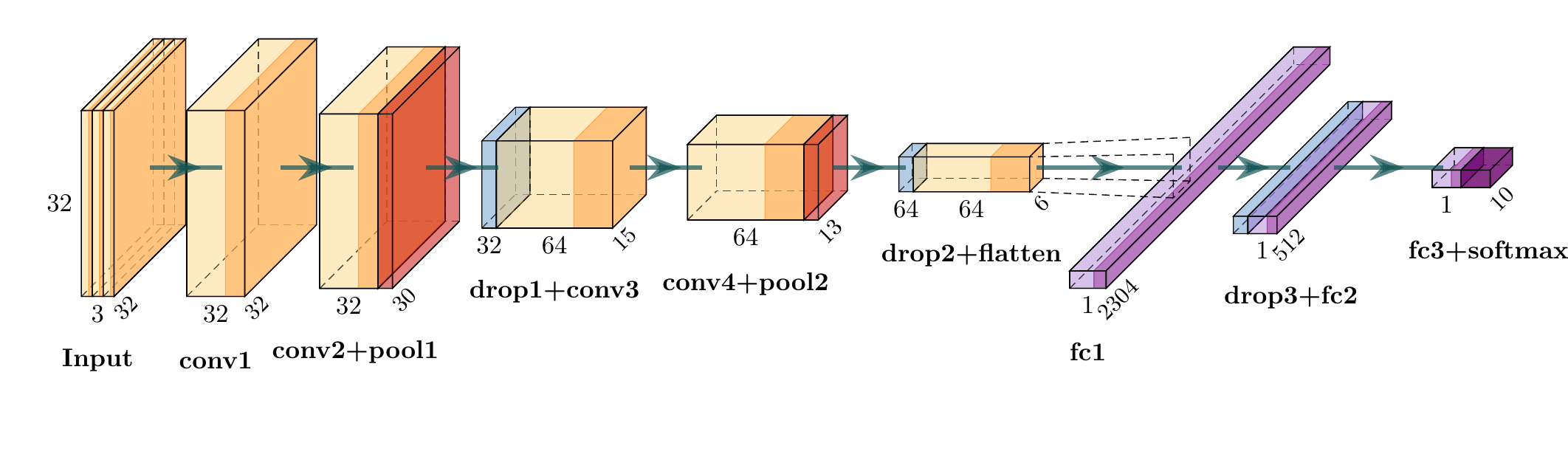}
    \caption{Convolutional neural network architecture applied in our tests on CIFAR-10. Each convolution is followed by a ReLU (rectified linear unit) activation. The dropout rate is 0.25 for the first two dropout layers and 0.50 for the third one, respectively.}
    \label{fig:cnn_cifar10}
\end{figure}

\commentMR{In order to test the general concept of MetaFusion, we perform experiments with} CIFAR-10~\cite{Krizhevsky2012}. The dataset is commonly used for image classification and contains 60k \commentMR{color} images of resolution $32\times32$ pixels in 10 classes, each class having \commentMR{the} same amount of samples. The CNN architecture we use in our experiments for this task (\cref{fig:cnn_cifar10}) is \commentMR{adopted} from \commentMR{the} Keras documentation~\cite{keras2015}, \commentMR{the network} is reported to achieve a validation accuracy of $79\%$ after 50 epochs of training.

\commentRC{To evaluate our method, we construct a rare class setup with ten CNNs. In this setup, the $i$-th CNN is trained on a CIFAR-10 (training data) subset assembled by randomly leaving out 90\% of the samples of class $i$.}
%
In~\cite{Buda2018} it already has been shown empirically that the application of the Maximum Likelihood decision rule positively affects the classification performance, not only increasing area under the receiver operating characteristic curve (AUROC) but also total accuracy. In particular, compensating for prior class probabilities increases the number of properly classified minority class samples. Based on this finding, we examine MetaFusion\commentMR{'s} behavior.

We evaluate \commentMR{the} ten CNNs on one and the same CIFAR-10 validation set consisting of 10k images with balanced class distribution. Each CNN is trained 50 epochs with categorical cross-entropy loss.
After adjusting the softmax probabilities by the priors (cf.~\cref{eq:ml-rule}) \commentMR{to perform the ML decision rule subsequently},
we derive three dispersion measures, namely entropy $E$, probability margin $M$ and variation ratio $V$. Note \commentMR{that}, since CIFAR-10 is an image classification task, \commentMR{the priors as well as the metrics are on the level of full images}
\commentRC{(e.g., consider $\mathcal{Z} = \{1\}$)}.
\commentMR{For each 
\commentRC{CNN},
the candidate images for MetaFusion are samples predicted by ML to belong to the \commentRC{trained} subset's minority class, \commentMR{but not by Bayes}}. As meta classifier we use GB that, based on $E,M,V$, classifies if an image is \commentMR{predicted} correctly or incorrectly. \commentRC{For GB we employ 15 boosting stages with maximum depth of 3 per tree and exponential loss function.}
\commentMR{In case an image's classification result is meta classified to be incorrect, we replace it with the class prediction obtained by the Bayes rule.}
\commentRC{MetaFusion is leave-one-out cross-validated.}

\begin{figure}[t]
    \centering
    \includegraphics[width=\linewidth]{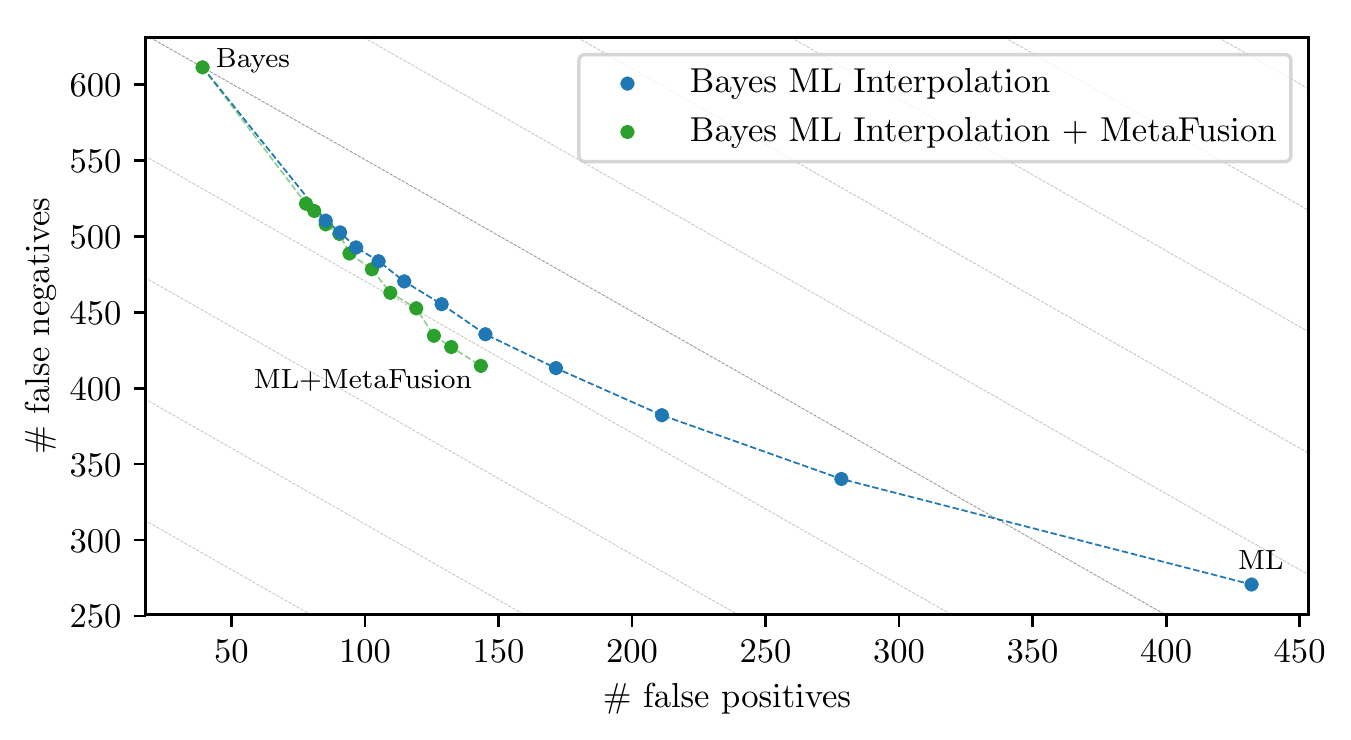}
    \caption{False-positives vs. false-negatives on CIFAR-10. The blue line is an interpolation between the Bayes and ML. Different points indicate equidistant interpolation degrees starting from 0.9, i.e., different degrees of decision thresholding. MetaFusion (green points) is applied in accordance to the interpolated class weightings. The diagonal lines denote level sets in which the sum of both errors equals the same value. 
    }
    \label{fig:error_cifar10}
\end{figure}

\commentMR{The main evaluation metrics that serve for our evaluation are the numbers of false-positives ($FP$) and false-negatives ($FN$) with respect to the minority class. In \cref{fig:error_cifar10} we see that, averaged over the ten CIFAR-10 subsets we obtain roughly 611 FNs and 39 FPs with Bayes whereas ML produces 271 FNs and 432 FPs. As a baseline we interpolate the priors between these two decision rules in order to understand how they translate into each other, i.e., we use the priors
\begin{equation}\label{eq:interpol}
    p_{z,\mathit{\alpha}}(y) = (1-\alpha) 1 + \alpha p_z(y) ~ \forall ~ y \in \mathcal{Y}, z \in \mathcal{Z},
\end{equation}
with $\alpha \in [0,1]$, resulting in the adjusted decision rule
\begin{equation}\label{eq:adj-decision-rule}
    d_\mathit{adj}(x,\alpha)_z := \argmax{y\in\mathcal{Y}} \frac{p_z(y|x)}{p_{z,\alpha}(y)}
\end{equation}
\commentRC{with $d_\mathit{adj}(x,0) = d_\mathit{Bayes}(x)$ and $d_\mathit{adj}(x,1) = d_\mathit{ML}(x)$. By v}arying the coefficient $\alpha$ we obtain the blue line in \cref{fig:error_cifar10} that may serve as an intuitive approach to balance FNs and FPs.
For each of the points given on the blue curve we apply MetaFusion (green line). Thus, many of the overproduced FPs are removed, however we also have to sacrifice some of the highly desired FNs at the same time. The diagonal gray lines visualizes level sets with respect to the sum of FNs and FPs which is the absolute number of errors, i.e., on each of these line the sum is constant. 
In our experiments, we choose equidistant interpolation degrees $\alpha$. Due to lack of data for MetaFusion the smaller $\alpha$ and/or the more confident the underlying image classification model, the smallest interpolation degree we use is $\alpha=0.9$. We notice that calibrating the class weightings leads to better average performance with respect to the sum of false-positives and false-negatives. We observe that applying MetaFusion reduces the sum of errors once more, nearly throughout all of investigated interpolation degrees (green line lying below blue line in~\cref{fig:error_cifar10}).
%
%
In order to further analyze this test, we state numbers for single runs in \cref{tab:cifar10_eval}, complemented with additional evaluation metrics.}
\begin{table}[t]
    \begin{center}
    {\caption{Performance comparison of Bayes, ML and MetaFusion on minority classes in CIFAR-10. In total, the performance of ten CNNs are reported, each CNN trained with a different minority class. To generate the unbalanced training dataset, $90\%$ of one class' samples were randomly removed.}\label{tab:cifar10_eval}}
    \vspace{.1in}
    \scalebox{0.75}{
    \begin{tabular}{c|ccc|cccc|cccc}
    \toprule
    \emph{}        & \multicolumn{3}{c}{Bayes} & \multicolumn{4}{|c}{Maximum Likelihood} & \multicolumn{4}{|c}{MetaFusion} \\
 $y$ & $F_{1}$ & $FP$    & $FN$    & $F_{1}$ & $FP$    & $FN$    & $\Delta$ & $F_{1}$ & $FP$    & $FN$    & $\Delta$ \\
 \midrule
0    & 0.66 & \textbf{85}     & 461     & 0.70 & 619     & \textbf{132}     & 1.62 & \textbf{0.72} & 175     & 332     & \textbf{0.70} \\
1    & 0.58 & \textbf{21}     & 585     & 0.78 & 169     & \textbf{253}     & 0.45 & \textbf{0.79} & 103     & 280     & \textbf{0.27} \\
2    & 0.41 & \textbf{39}     & 729     & \textbf{0.54} & 698     & \textbf{367}     & 1.82 & 0.50 & 112     & 629     & \textbf{0.73} \\
3    & 0.15 & \textbf{17}     & 915     & \textbf{0.53} & 502     & \textbf{452}     & 1.05 & 0.42 & 138     & 691     & \textbf{0.54} \\
4    & 0.46 & \textbf{70}     & 682     & 0.60 & 852     & \textbf{203}     & 1.63 & \textbf{0.63} & 195     & 452     & \textbf{0.54} \\
5    & 0.34 & \textbf{35}     & 787     & \textbf{0.61} & 528     & \textbf{330}     & 1.08 & 0.56 & 207     & 532     & \textbf{0.67} \\
6    & 0.67 & \textbf{41}     & 473     & 0.75 & 336     & \textbf{202}     & 1.09 & \textbf{0.77} & 137     & 293     & \textbf{0.53} \\
7    & 0.65 & \textbf{24}     & 506     & \textbf{0.74} & 248     & \textbf{261}     & 0.91 & 0.73 & 106     & 368     & \textbf{0.60} \\
8    & 0.69 & \textbf{29}     & 459     & 0.79 & 183     & \textbf{234}     & 0.68 & \textbf{0.79} & 126     & 260     & \textbf{0.49} \\
9    & 0.64 & \textbf{31}     & 516     & \textbf{0.76} & 185     & \textbf{273}     & 0.63 & 0.76 & 132     & 309     & \textbf{0.49} \\
    \midrule
$\bar{y}$ & 0.52  & \textbf{39}      & 611     & \textbf{0.68}  & 432     & \textbf{271}     & 1.09 & 0.67  & 143     & 415     & \textbf{0.56} \\
    \midrule
     & \multicolumn{11}{c}{Averaged total accuracy score on validation set  } \\
    \midrule
    $\bar{y}$ & \multicolumn{3}{c|}{0.75451} & \multicolumn{4}{c|}{\textbf{0.76563}} & \multicolumn{4}{c}{0.76707} \\
    \bottomrule
    \end{tabular}
    }
    \end{center}
\end{table}

\commentMR{As an overall performance measure, although being skewed towards majority classes, we report the score $F_1 = 2TP / (2TP+FP+TP)$ with $TP$ being the number of true-positives.
Another measure for MetaFusion is the ratio between prediction errors. For any decision rule $d_{adj}:[0,1]^{|\mathcal{Y}|} \times \mathbb{R} \mapsto \mathcal{Y}$, the slope
\begin{equation} \label{eq:delta}
    \Delta(d_\mathit{adj}) = \frac{FP(d_\mathit{adj})-FP(d_\mathit{Bayes})}{FN(d_\mathit{Bayes})-FN(d_\mathit{adj})}
\end{equation}
with $d_\mathit{adj}$ such that $\commentRC{FN(d_\mathit{Bayes})-FN(d_{adj}) \neq 0}$ describes how many additional FPs we have to accept for removing a single FN compared to the Bayes decision rule. The smaller $\Delta$, the \commentMR{more favorable} the trade-off between the two error rates. In fact, $\Delta < 1$ indicates that for the considered minority class 
the total number of errors is decreased by $d_\mathit{adj}$ compared to $d_\mathit{Bayes}$ (whereas it may increase for the other classes).
}

\commentMR{The average $F_1$ score is $67\%$, marginally less than with ML ($68\%$). This outcome is mainly caused by the class 3 \commentRC{(cat)} where ML considerably outperforms MetaFusion by $11$ percent points. For the remaining classes the exchange is at most 5 percent points with either MetaFusion achieving a higher score than ML or vice versa. MetaFusion is superior to ML in average $\Delta$ by taking roughly only one FP in order to reduce two FNs. Hence, with respect to average error rates, MetaFusion outperforms Bayes and ML.
On average $\Delta(d_\mathit{ML})=1.09$, i.e., ML produces slightly more than one FP to reduce one FN in comparison to Bayes. Moreover, also compared to Bayes, the amount of predicted minority class instances is significantly increased, leading to an improved performance per minority class by $16\%$ on average and also in total accuracy by $1.11\%$. This result confirms the finding from~\cite{Buda2018}.}
%
\commentMR{Summarizing this test}, we have shown empirically that the basic concept of MetaFusion works for image classification by implementing a minimal version of the method and reporting numerical results. Next, we aim at extending our method to the application-relevant and more complex task of semantic segmentation.

\section{NUMERICAL RESULTS FOR CITYSCAPES}\label{sec:cityscapes}
\commentRC{%
Semantic segmentation is a crucial step in the process of perceiving a vehicle's surroundings for automated driving. Therefore, we perform tests on the Cityscapes dataset~\cite{Cordts2016Cityscapes} which consists of 2,975 pixel-annotated street scene images of resolution $2048 \times 1024$ pixels used for training and further 500 images for validation purposes. CNNs can be trained either on 19 classes or 8 aggregated coarse categories. Our main focus lies in avoiding non-detected humans (ideally without producing any false-positive predictions). As all images are recorded in urban street scenes (thus naturally boosting the occurrence \commentMR{of} persons), classes like wall, fence or pole are as rare as pedestrians in terms of pixel frequency in the dataset. This would lead to class priors, when estimating via pixel-wise frequency, conflicting with human common sense due to the possible preference of static objects over persons. Therefore, we use category priors treating objects more superficially (\commentMR{by aggregrating all clases into the 8 predefined categories}), with pedestrians and rider aggregated to ``\emph{human}'' class then being significantly underrepresented relative to all remaining categories.
}

We perform the Cityscapes experiments using DeeplabV3+ networks~\cite{Chen2018ECCV} with MobileNetV2~\cite{Sandler2018} and Xception65~\cite{Chollet2017} backbones. We apply MetaFusion per predicted human segment as presented in~\cref{sec:metafusion} and evaluate only the human class in the Cityscapes validation data. \commentRC{As meta classifier we employ GB with 27 boosting stages, maximum depth of 3 per tree, exponential loss and 5 features to consider when looking for the best split. MetaFusion is 5-fold cross-validated.} Numerical results are listed in~\cref{tab:cityscapes_eval}.

\begin{table}[t]
    \begin{center}
    \caption{Performance comparison of different decision rules and MetaFusion for DeeplabV3+ with MobileNetV2~\cite{Sandler2018} and Xception65 backbones on Cityscapes. Different adjusted decision rules are obtained according to~\cref{eq:adj-decision-rule}.}\label{tab:cityscapes_eval}
    \vspace{.1in}
    \scalebox{0.75}{
    \begin{tabular}{c|cccc|cccc}
    \toprule
Priors interpol.        & \multicolumn{4}{|c}{Adjusted Decison Rule} & \multicolumn{4}{|c}{MetaFusion} \\
  degree $\alpha$ & $\mathit{mIoU}$ & $FP$    & $FN$    & $\Delta$ & $\mathit{mIoU}$ & $FP$    & $FN$    & $\Delta$ \\
 \midrule
     & \multicolumn{8}{c}{DeeplabV3+ MobileNetV2 on Cityscapes validation set} \\
 \midrule
0.000 (Bayes)   & 0.684  & 865     & 839     & - & 0.684  & 865     & 839     & - \\
\midrule
0.900   & 0.675  & 1644     & \textbf{631}     & 3.735 & 0.683 & \textbf{1167}     & 720     & \textbf{2.538} \\
0.950   & 0.668  & 1988     & \textbf{571}     & 4.190 & 0.682 & \textbf{1169}     & 670     & \textbf{1.799} \\
0.975   & 0.661  & 2352     & \textbf{533}     & 4.860 & 0.681 & \textbf{1191}     & 648     & \textbf{1.701} \\
0.990   & 0.653  & 2827     & \textbf{496}     & 5.720 & 0.680 & \textbf{1247}     & 611     & \textbf{1.676} \\
0.995   & 0.649  & 3155     & \textbf{485}     & 6.469 & 0.680 & \textbf{1329}     & 586     & \textbf{1.834} \\
1.000 (ML)   & 0.600  & 4885     & \textbf{476}     & 11.074 & 0.680 & \textbf{1606}     & 553     & \textbf{2.590} \\
     \midrule
     & \multicolumn{8}{c}{DeeplabV3+ Xception on Cityscapes validation set} \\
 \midrule
0.000 (Bayes)   & 0.753  & 774      & 679     & - & 0.753  & 774     & 679     & - \\
\midrule
0.900           & 0.746  & 1314     & \textbf{530}     & \textbf{3.624} & 0.752 & \textbf{1055}     & 614     & 4.323 \\
0.950           & 0.742  & 1579     & \textbf{487}     & 4.193 & 0.752 & \textbf{1079}     & 583     & \textbf{3.177} \\
0.975           & 0.737  & 1783     & \textbf{458}     & 4.566 & 0.751 & \textbf{1118}     & 571     & \textbf{3.185} \\
0.990           & 0.732  & 2068     & \textbf{433}     & 5.260 & 0.751 & \textbf{1103}     & 549     & \textbf{2.531} \\
0.995           & 0.731  & 2219     & \textbf{425}     & 5.689 & 0.750 & \textbf{1154}     & 532     & \textbf{2.585} \\
1.000 (ML)      & 0.705  & 3003     & \textbf{421}     & 8.640 & 0.750 & \textbf{1272}     & 508     & \textbf{2.912} \\
    \bottomrule
    \end{tabular}
    }
    \end{center}
\end{table}
\begin{figure}[t]
    \centering
\includegraphics[width=\linewidth]{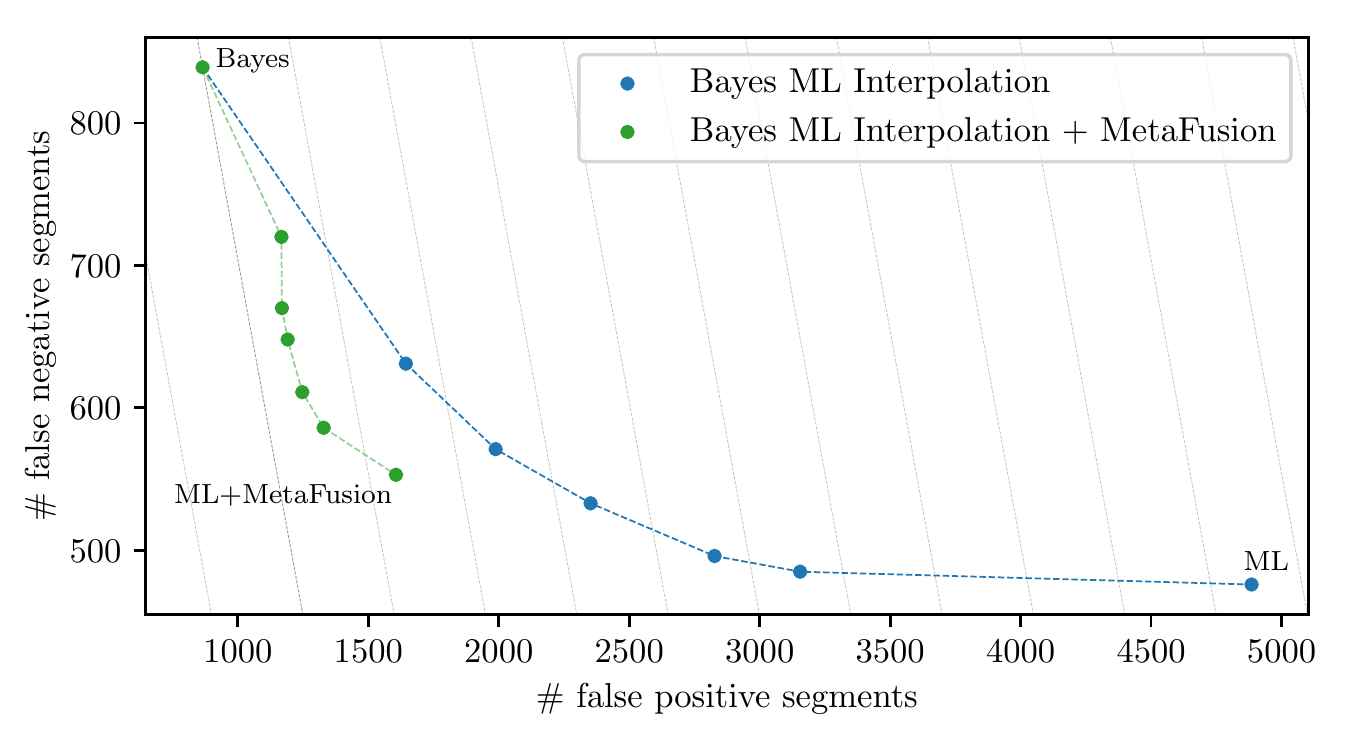}
\caption{False-positives vs. false-negatives of person segments for MobileNetV2 on Cityscapes. The diagonal lines denote level sets in which the sum of both errors equals the same value.}
    \label{fig:error_mn}
\end{figure}
\begin{figure}[t]
    \centering
\includegraphics[width=\linewidth]{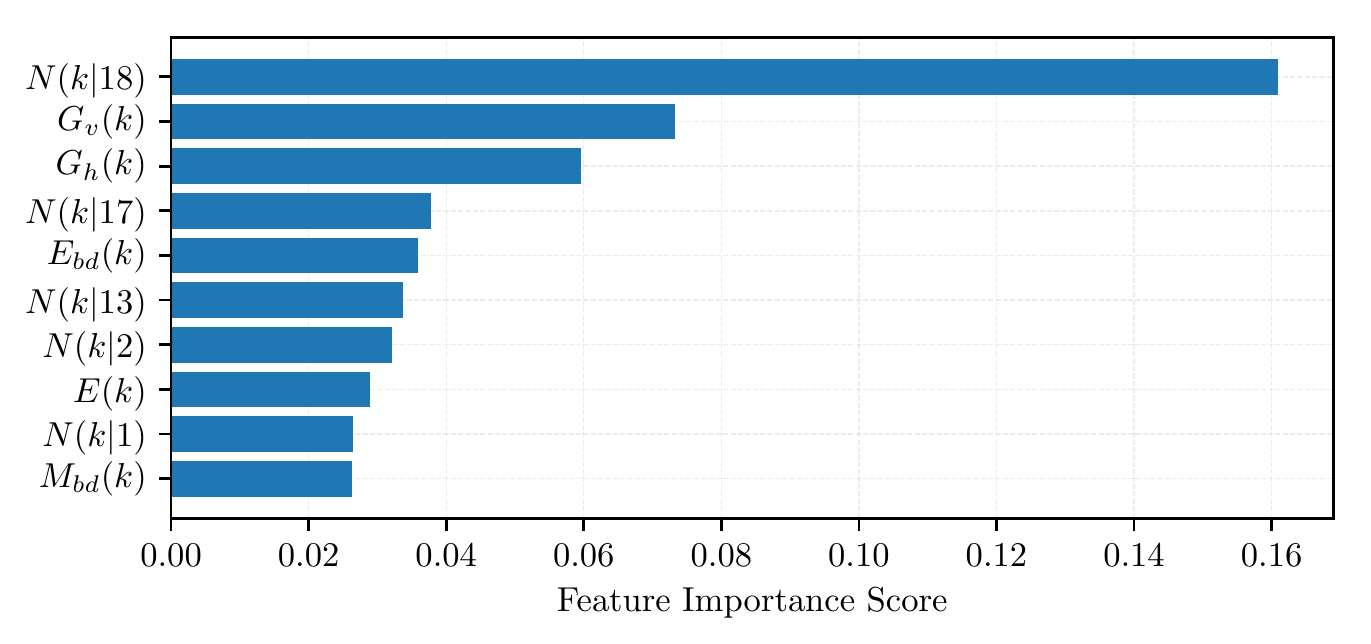}
\caption{Feature importance scores of the gradient-boosting classifier for MobileNetV2 applied on all disjoint ML and Bayes human segments. The score is averaged over all random cross-validation splits and only the ten features with the highest score are depicted. In total we used 56 metrics as meta model input. $N$ and $G$ are defined in~\cref{sec:prediction-error-classification}. $E$ and $M$ denotes the segment-wise averaged entropy and probability margin, respectively, with $bd$ indicating the restriction on the segment's boundary.}
    \label{fig:importance_mn}
\end{figure}

\begin{figure*}
    \begin{center}
    \input{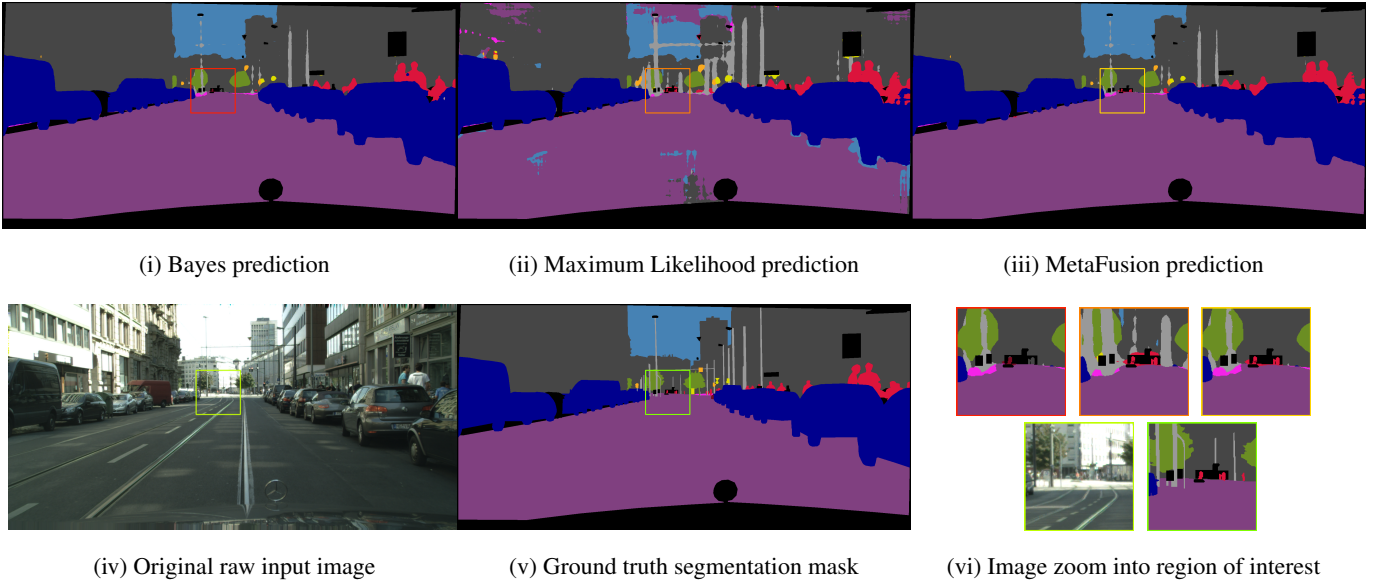}
    \end{center}
    \caption{Example of generated segmentation masks with MobileNetV2. In the top row: prediction mask using Bayes (i), ML (ii) and MetaFusion (iii). In the bottom row: raw input image (iv), corresponding annotated ground truth mask (v) and zoomed views into the region of interest marked in the latter images (vi). By comparing the prediction masks, we observe couple of person segments (red color) for which the decision rules disagree and which are correctly identified as false-positive, according to the ground truth, using MetaSeg. In the end, with MetaFusion we obtain a segmentation mask similar at large to the standard Bayes mask but with some additionally detected person instances that are rather small and barely visible in the original image.}
\end{figure*}

\commentRC{
Similar to the experiments in~\cref{sec:cifar10}, we interpolate between Bayes and ML priors according to~\cref{eq:interpol} but now for every pixel location $z \in \mathcal{Z}$. We again observe that an interpolation degree of $\alpha<0.9$ for the adjusted decision rules (see~\cref{eq:adj-decision-rule}) leads to a lack of meta training data. Moreover, we choose unevenly spaced steps $\alpha \in \{0.9, 0.95, 0.975, 0.99, 0.995, 1 \}$ due to a drastic increase in error rates the bigger the interpolation degree.}

\commentRC{
For MobileNetV2, see also~\cref{fig:error_mn}, we observe that the number of false-positives increases from 865 up to 4885 when applying ML instead of Bayes while the number of false-negatives decreases from 839 down to 476. This results in large $\Delta=11.07$ expressing that roughly 11 FPs are produced in order to remove one single FN. Clearly, there is an overproduction of predicted human segments \commentMR{that} we want to keep under control using MetaFusion.}

\commentRC{
By applying MetaFusion, the number of false-positives is reduced to a third of ML false-positives while keeping more than two thirds (78.79\%) of additional true-positives. This results in $\Delta=2.59$ \commentMR{which} is a significant decrease \commentMR{compared to plain ML without MetaFusion}. With respect to the overall performance, measured by \emph{mean} IoU, MetaFusion sacrifices 0.4\% and ML 8.4\% for detecting the false-negatives additionally to Bayes. \commentMR{In our experiments we observe that our approach works better the more segments are available for which the decision rules disagree}. Therefore, the performance gain with respect to \commentMR{the} total number of errors is most significant for $\alpha=1.0$. For decreasing interpolation degrees, we observe a successive reduction of total number errors for the adjusted decision rules. Different to the findings for CIFAR-10, the class weightings\commentMR{'} adjustment does not lead to a better performance than Bayes with respect to the absolute number of errors. \commentRC{However, when avoiding FNs is considered to be more important than FPs, our method proposes alternative decision rules that are more attractive than plain decision rules for a large set of error weightings.}}
%
Just like for CIFAR-10, for every investigated $\alpha$ MetaFusion is superior to ML regarding the failure trade-off $\Delta$ producing 1.68 additional FPs for removing one single FN as its best performance. In addition, we can conclude that our approach outperforms probability thresholding with respect to the error rates on human segments.

\commentRC{For the stronger DeeplabV3+ model with Xception65 network backbone, we observe similar effects in general. Compared to MobileNetV2, MetaFusion's performance gain over adjusted decision rules is not as great. This is primarily due to the higher confidence scores in the softmax output of the underlying CNN. They prevent the adjusted decision rules from producing segments for which the decision rules disagree. Therefore, the training set size for the meta classifier is rather small even resulting in a worse $\Delta$ for MetaFusion than for the adjusted decision rule when $\alpha=0.90$. Nevertheless, the latter does not hold for the remaining investigated interpolation degrees. Indeed, MetaFusion accepts in average 2.8 FPs for removing one single FN which is more than half of the average $\Delta$ for the adjusted decision rules.}


\commentMR{In order to find out which of the constructed metrics contribute most to meta classification performance, we analyze our trained GB with respect to feature importance.}
The 
\commentMR{latter} is a measure indicating the relative importance of each feature variable in a GB model. In a decision tree the importance is computed as
\begin{equation}
    I_n(t) = n(t) Q(t) -  n_{\mathit{left}}(t) Q_{\mathit{left}}(t) - n_{\mathit{right}}(t) Q_{\mathit{right}}(t)
\end{equation}
with $Q(t)$ the Gini impurity~\cite{hastie2009} and $n(t)$ the weighted number of samples in node $t \in \mathcal{T}$ (the weighting corresponds to the portion of all samples reaching node $t$). Moreover, by $\mathit{left}$ and $\mathit{right}$ we denote the respective children nodes.
Then the importance of $\hat{f}$ of feature / uncertainty metric $m \in [0,1]$ is computed as

\begin{equation}
    I(m) = \frac{ \displaystyle \sum_{t \in \mathcal{T}} \chi(t|m) I_n(t)}{ \displaystyle \sum\limits_{t \in \mathcal{T}} I_n(t)}
\end{equation}
with 

\begin{equation}
    \chi(t|m) = 
    \begin{cases}
    1, ~ \text{if node $t$ splits on feature $m$} \\
    0, ~ \text{else}
    \end{cases} ~ .
\end{equation}
The ten features of highest importance (\commentMR{in experiments with MobileNetV2}) are reported in~\cref{fig:importance_mn}. By \commentMR{a} large margin, a segment's neighborhood including class id 18, which \commentMR{corresponds to} bicycles, has the strongest effect on GB. This is plausible since
\commentRC{a bicycle segment adjacent to a human segment can be viewed as an indicator that this human segment is indeed present, i.e., a true-positive.}
Having less than half the importance score, the geometric center still has \commentMR{a} relatively high impact \commentMR{on} GB. We notice that ML produces many (false-positive) segments close to the image borders. This is a consequence of applying pixel-wise ML which GB takes into account. The dispersion measures entropy and probability margin are considered as important features as well expressing the CNN's uncertainty about its prediction. In~\cite{rottmann2018}, it already has been shown that these two metrics are well-correlated to the segment-wise IoU. GB also uses these correlation\commentMR{s} to perform the meta classification. \commentMR{In contrast to the findings in \cite{rottmann2018}, dispersion measures at segment} boundaries have greater impact than the dispersion of the interior. This high uncertainty at the boundaries can be interpreted as disturbances for class predictions in a segment's surrounding and may indicate that the investigated segment is a false-positive. 
Moreover, the remaining features in the top ten of highest importance are neighborhood \commentMR{statistics for the classes} (in descending order) motocycle, car, building and sidewalk.


\section{CONCLUSION}
In this work, we presented a novel post-processing approach for semantic segmentation. As minority classes are often of highest interest in many real-world applications, the non-detection of their instances might lead to fatal situations and therefore must be treated carefully. In particular, the class person is one minority class in street scene datasets. We compensate unbalanced class distributions by applying the Maximum Likelihood decision rule that detects a significantly larger number of humans, but also causes an overproduction of false-positive predictions of the same class. By deriving uncertainty measures per predicted segment and passing them through a gradient-boosting classifier, we are able to detect false-positive segment predictions in the ML mask in an automated and computationally cheap fashion. We remove these segments which are identified as incorrect and replace them with the Bayes mask. 
\commentMR{In this way, we significantly reduce the number of false-positives, at the same time only sacrificing a small number of detected false-negatives and also only resulting in a minor overall performance loss in comparison to the standard Bayes decision rule} in the Cityscapes dataset.
In fact, our method, which we term ``\emph{MetaFusion}'', outperforms decision rules with different class weightings obtained by interpolating between Bayes and ML rule, i.e., MetaFusion outperforms pure probability thresholding with respect to both error rates, false-positive and false-negative, of class human. This result holds for the investigated DeeplabV3+ models with MobileNetV2 and Xception65 backbones whereby the performance gain is more substantial the greater the difference between the Bayes and ML mask. Furthermore, we tested the basic concept of MetaFusion on an image classification problem as well. Although we applied only a minimal version on the CIFAR-10 dataset, we observed similar results to Cityscapes demonstrating the method's generalization capabilities for various tasks.
\commentMR{MetaFusion can be viewed as a general concept for trading improved false-positive detection for additional performance on rare classes.}

For future work we plan to improve our meta classification approach with further heatmaps, metrics as well as component-sensitive to time dynamics. Our approach might also be suitable to serve for query strategies in active learning.
Our source code for reproducing experiments is publicly available on GitHub, see \url{https://github.com/robin-chan}.


\vspace{1.8ex}
\ack This work is in part funded by Volkswagen Group Innovation. We thank Jan David Schneider and Matthias Fahrland for fruitful discussions and 
programming support.

\newpage
\bibliography{ecai}
\end{document}